%% file: Main.tex
\documentclass{article}

\usepackage[preprint]{neurips_2020}

\usepackage{graphicx}
\usepackage{caption}
\usepackage{subcaption}
\usepackage[utf8]{inputenc} 
\usepackage[T1]{fontenc}    
\usepackage[draft]{hyperref}       
\usepackage{url}            
\usepackage{booktabs}       
\usepackage{amsfonts}       
\usepackage{nicefrac}       
\usepackage{tikz}
\usepackage{tabularx,booktabs}
\usepackage{amsmath,amssymb,amsfonts}

\title{Self-Competitive Neural Networks}

\author{%
	Iman Saberi \\
	Department of Electrical and Computer Engineering\\
	University of Tehran\\
	Tehran, Iran \\
	\texttt{iman.saberi@ut.ac.ir} \\
	\And
	Fathiyeh Faghih \\
	Department of Electrical and Computer Engineering\\
	University of Tehran\\
	Tehran, Iran \\
	\texttt{f.faghih@ut.ac.ir} \\
}

\begin{document}
	
	\maketitle

\input{Abstract}

\input{Introduction}

	\input{RelatedWorks}

\input{SelfCompetetiveNeuralNetworks}

\input{ExperimentalResults}

\input{Conclusion}

	\bibliography{ref}
\bibliographystyle{apalike}
\end{document}

%% file: Abstract.tex
\begin{abstract}
	
	Deep Neural Networks (DNNs) have improved the accuracy of classification problems in lots of applications. One of the challenges in training a DNN is its need to be fed by an enriched dataset to increase its accuracy and avoid it suffering from overfitting. One way to improve the generalization of DNNs
	is to augment the training data with new synthesized adversarial samples. Recently, researchers have worked extensively to propose methods for data augmentation. In this paper, we generate adversarial samples to refine the Domains of Attraction (DoAs) of each class. In this approach, at each stage, we use the model learned by the primary and generated adversarial data (up to that stage) to manipulate the primary data in a way that look complicated to the DNN. The DNN is then retrained using the augmented data and then it again generates adversarial data that are hard to predict for itself. As the DNN tries to improve its accuracy by competing with itself (generating hard samples and then learning them), the technique is called Self-Competitive Neural Network (SCNN). To generate such samples, we pose the problem as an optimization task, where the network weights are fixed and use a gradient descent based method to synthesize adversarial samples that are on the boundary of their true labels and the nearest wrong labels. Our experimental results show that data augmentation using SCNNs can significantly increase the accuracy of the original network. As an example, we can mention improving the accuracy of a CNN trained with 1000 limited training data of MNIST dataset from 94.26\% to 98.25\%.

\end{abstract}

%% file: Introduction.tex
\section{Introduction}
Deep learning models have performed remarkably well on many classification problems. With the advent of Convolutional Neural Networks (CNNs)~\cite{lecun1998gradient}, significant improvements have been reported in computer vision tasks. One of the main challenges in training a deep neural model is providing a big dataset in order to prevent model from overfitting. The challenge is more significant in small datasets, such as medical image analysis. Data augmentation is a known solution in the literature to improve model generalization. 

One of the well-known techniques for data augmentation are Generative Adversarial Networks (GANs)~\cite{goodfellow2014generative}. They are used to generate new data in order to inflate the training dataset~\cite{nielsen2019gan,goodfellow2014generative,bowles2018gan,antoniou2017data}. There are, however, challenges in data generation using these networks. First, the aim of GANs is to find a Nash equilibrium of a non-convex game with continuous and high-dimensional parameters, while they are typically trained based on a gradient descent based technique designed to minimize a cost function (instead of finding a Nash equilibrium of a game). Therefore, these algorithms may fail to converge~\cite{Goodfellow2016}, and hence, synthesis of high resolution data may be very difficult using this technique. Another limitation of GANs is their need to substantial amount of primary data in order to train the discriminator well, and hence, they are not practical in small datasets~\cite{Shorten2019}.

In this paper, we propose a novel approach in this field that concentrates on the weaknesses of the functional structure of DNNs in order to improve their accuracy.  The non-polynomial architecture of DNNs consists of deep linear and nonlinear operations, and hence, the Domains of Attraction (DoAs) of each class  cannot be easily determined. We present a method to refine the DoAs of the network by synthesizing harder samples from the primary input dataset.
 The synthesizer network modifies each sample in a way that it is  located on the boundary of its true label and its nearest wrong label in the trained embedded network. The synthesizer tries to generate more complicated samples from the primary data, and then the embedded network tries to learn them correctly. This cycle is repeated as many times as the accuracy of the embedded network increases.
Similar to GANs, our approach consists of two networks, a synthesizer and an embedded network. However,  unlike GANs that try to play a minimax  game and converge to a Nash equilibrium, in our approach, the two networks separately try to minimize their loss function based on a gradient descent method in order to improve the accuracy and robustness of the embedded network.

Our experimental results demonstrate that our proposed technique can improve the accuracy of networks, especially in small datasets. We selected  a limited set of 1000 training data from the MNIST dataset, and fed them to an SCNN. We observed that the accuracy of the baseline well-trained embedded network was increased from 94.26\% to 98.25\% using our technique. We also did experiments on Fashoin MNIST and Cifar10 datasets as harder datasets. The results show 1.35\% increase in the accuracy of the Fashion MNIST dataset and 4.85\% increase in the accuracy of the Cifar10 dataset, compared to the baseline Resnet18 Model.



%% file: RelatedWorks.tex
\section{Related Work}

Mining hard examples was previously studied in the literature~\cite{uijlings2013selective,shrivastava2016training}. The idea is to select or generate optimal and informative samples in order to enrich the dataset.
Data augmentation techniques in the literature can be categorized into data wrapping and oversampling methods. Data wrapping augmentations transform the existing samples, such that their labels are preserved. Geometric and color transformations, random erasing, adversarial training, and style transfer networks are examples of data wrapping techniques. Oversampling augmentation methods generate new instances and add them to the training set. Oversampling encompasses augmentations, such as mixing images, feature space augmentation, and GANs. These two categories do not form a mutually exclusive dichotomy~\cite{Shorten2019}. Our approach is a data wrapping augmentation technique that tries to manipulate each sample, such that it is located on the boundary of its true label and its nearest wrong label.

In~\cite{fawzi2016adaptive}, the authors propose a method to seek small transformations that yield maximal classification loss on the transformed sample based on a trust region strategy. This work is similar to our idea in the criterion of generating informative augmented data. However, our strategy in synthesizing augmented samples is totally different. The proposed algorithm in~\cite{fawzi2016adaptive} selects a set of possible transformations, where each one has a specific degree of freedom. The algorithm applies a set of transformations that make the cost function have the most value. Our technique is different in that a gradient descent method tries to move each sample in a direction to be located on the decision boundary for that sample.

The closest work to this paper is~\cite{peng2018jointly}, where the authors designed an augmentation network that competes against a target network by generating hard examples (using GAN structure). The algorithm selects a set of augmentations that have the maximal loss against random augmentation samples. The idea is to apply a reward/penalty strategy and formalize the problem as a minimax game for effective generation of hard samples. Our paper is different in that we do not make the generator network to select the best strategy from a set of predefined strategies. Instead, a gradient descent method decides what is the best transformation parameters in order to manipulate each sample. Also, we pose the problem as minimizing two cost functions separately, instead of playing a minimax game.

%% file: SelfCompetetiveNeuralNetworks.tex
\section{Self-Competitive Neural Network}

	\subsection{SCNN Architecture}
	The architecture of a Self-Competitive Neural Network (SCNN) contains two main parts: 
	\begin{enumerate}
		\item An embedded neural network:
			The goal of SCNN is to improve the accuracy of this network. The internal design of this network is crucial in order to have a powerful model. The embedded neural network can be of any type, such as Convolutional Neural Network (CNN).

		\item An Adversarial Data Synthesizer (ADS) network:
		 	This network is constructed by  concatenation of a number of differentiable components and the embedded network. The aim of this network is synthesizing difficult data for the embedded network. The differentiable components take the input data and try to learn their parameters in a way that the synthesized data is predicted as the nearest wrong label by the embedded network.
			The details of the internal structure of this component will be discussed in Section~\ref{section:main}.
	\end{enumerate}

\begin{figure}
	\centering
	\scalebox{.8}{
		\begin{tikzpicture}[scale=0.5, every node/.style={scale=0.5},x=0.75pt,y=0.75pt,yscale=-1,xscale=1]
		
		\draw  [color={rgb, 255:red, 0; green, 0; blue, 0 }  ,draw opacity=0 ][fill={rgb, 255:red, 145; green, 222; blue, 60 }  ,fill opacity=0.58 ] (269,134) .. controls (269,93.68) and (301.68,61) .. (342,61) .. controls (382.32,61) and (415,93.68) .. (415,134) .. controls (415,174.32) and (382.32,207) .. (342,207) .. controls (301.68,207) and (269,174.32) .. (269,134) -- cycle ;
		\draw  [color={rgb, 255:red, 0; green, 0; blue, 0 }  ,draw opacity=0 ][fill={rgb, 255:red, 140; green, 189; blue, 246 }  ,fill opacity=0.71 ] (437,316) .. controls (437,275.68) and (469.68,243) .. (510,243) .. controls (550.32,243) and (583,275.68) .. (583,316) .. controls (583,356.32) and (550.32,389) .. (510,389) .. controls (469.68,389) and (437,356.32) .. (437,316) -- cycle ;
		\draw  [color={rgb, 255:red, 0; green, 0; blue, 0 }  ,draw opacity=0 ][fill={rgb, 255:red, 245; green, 166; blue, 35 }  ,fill opacity=0.55 ] (99,318) .. controls (99,277.68) and (131.68,245) .. (172,245) .. controls (212.32,245) and (245,277.68) .. (245,318) .. controls (245,358.32) and (212.32,391) .. (172,391) .. controls (131.68,391) and (99,358.32) .. (99,318) -- cycle ;
		\draw    (215.88,260.01) -- (278.91,188.75)(218.12,261.99) -- (281.16,190.73) ;
		\draw [shift={(286,183)}, rotate = 491.5] [fill={rgb, 255:red, 0; green, 0; blue, 0 }  ][line width=0.08]  [draw opacity=0] (10.72,-5.15) -- (0,0) -- (10.72,5.15) -- (7.12,0) -- cycle    ;
		\draw    (400.15,180.03) -- (460.33,251.16)(397.85,181.97) -- (458.04,253.1) ;
		\draw [shift={(465,259)}, rotate = 229.76] [fill={rgb, 255:red, 0; green, 0; blue, 0 }  ][line width=0.08]  [draw opacity=0] (10.72,-5.15) -- (0,0) -- (10.72,5.15) -- (7.12,0) -- cycle    ;
		\draw    (437.02,317.5) -- (254.02,319.41)(436.98,314.5) -- (253.98,316.41) ;
		\draw [shift={(245,318)}, rotate = 359.4] [fill={rgb, 255:red, 0; green, 0; blue, 0 }  ][line width=0.08]  [draw opacity=0] (10.72,-5.15) -- (0,0) -- (10.72,5.15) -- (7.12,0) -- cycle    ;
		
		\draw (284,105) node [anchor=north west][inner sep=0.75pt]   [align=left] {\begin{minipage}[lt]{80.37804000000001pt}\setlength\topsep{0pt}
			\begin{center}
			\textbf{Phase 1:}\\\textbf{training with primary data}
			\end{center}
			
			\end{minipage}};
		\draw (449,285) node [anchor=north west][inner sep=0.75pt]   [align=left] {\begin{minipage}[lt]{83.23608000000002pt}\setlength\topsep{0pt}
			\begin{center}
			\textbf{Phase 2:}\\\textbf{generating}\\\textbf{ adversarial data}
			\end{center}
			
			\end{minipage}};
		\draw (110,285) node [anchor=north west][inner sep=0.75pt]   [align=left] {\begin{minipage}[lt]{83.23608000000002pt}\setlength\topsep{0pt}
			\begin{center}
			\textbf{Phase 3:}\\\textbf{training with}\\\textbf{ adversarial data}
			\end{center}
			
			\end{minipage}};
		\draw (283,230) node [anchor=north west][inner sep=0.75pt]   [align=left] {\begin{minipage}[lt]{85.47804000000001pt}\setlength\topsep{0pt}
			\begin{center}
			\textbf{Self Competitive }\\\textbf{Neural Networks}\\\textbf{(SCNN)}
			\end{center}
			
			\end{minipage}};

		\end{tikzpicture}
	}
	\caption{The life cycle of training a SCNN}
	\label{phases}
\end{figure}
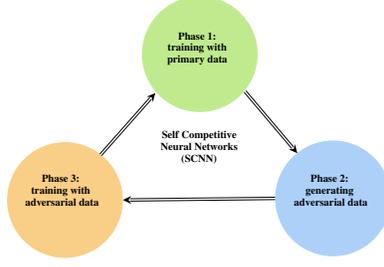

	The training life cycle of an SCNN has three main phases (Fig.~\ref{phases}): 
	\begin{enumerate}
		\item Training with primary data: In this phase, the embedded network is trained by the main input dataset.
		\item Adversarial data synthesis: After training the embedded network by the primary data, the weights of the embedded network are set as immutable, and  the primary input data is manipulated in a way that the  embedded network predicts the synthesized data by a wrong label. In other words, the ADS network manipulates the input data in a way that they are more difficult for the embedded network to predict correctly.
		\item Training with adversarial data:  In this phase, the SCNN uses the synthesized data from the previous phase,  sets the weights of the embedded network as mutable, and the embedded network is trained by the synthesized adversarial data. 
	\end{enumerate}
	
	You can see the embedded network in the both the training phases and the data generation phase. The difference is that in the training phases, the weights of the embedded network are mutable and learned during these steps. However, in the data generation phase, the weights are an immutable part of the ADS network, and hence, they are not changed.
	
	The three phases are repeated in a cycle as many times as the accuracy of the embedded network improves. In each cycle, the weights of the embedded network are tuned by the primary data (phase 1), the ADS synthesizes new data from the primary input data that seem harder for the embedded network to predict correctly (phase 2), and the embedded network is trained by the synthesized harder boundary data (phase 3).
	This procedure can be considered as {\em{online}} strategy, where the underlying embedded network is improved gradually by harder data.
	 Another approach is {\em{offline}}, where the model is trained from scratch by the primary and harder data generated in the training life cycle of an SCNN. It is obvious that the offline approach is more expensive, but as we will show in Section~\ref{section:results}, the resulting accuracy of this approach is better than the online strategy.

\section{Details of SCNN}

\label{section:main}
	The main idea of  SCNN
can be thought of as a  competition between the embedded neural network and the adversarial data synthesizer, where during the training phases (both  training with primary data and adversarial data), the embedded network tries to correctly predict the labels of the primary and adversarial data. On the other hand, during the adversarial data synthesis, the SCNN tries to manipulate the input data in a way that the embedded network predicts the manipulated data with wrong labels.


%
	
	For simplicity in explanation and without loss of generality, consider a MultiLayer Perceptron (MLP) as an embedded network. A loss function is defined in a way that explains how much the outputs of the network are different from the correct labels. The optimizer uses gradient descent to minimize this function. 
%
%
%

\begin{figure}[htbp]
	\centering
	\begin{tikzpicture}[scale=0.5, every node/.style={scale=0.5},x=0.75pt,y=0.75pt,yscale=-1,xscale=1]

\draw   (376,158.32) .. controls (376,146.8) and (385.33,137.47) .. (396.85,137.47) .. controls (408.36,137.47) and (417.7,146.8) .. (417.7,158.32) .. controls (417.7,169.83) and (408.36,179.17) .. (396.85,179.17) .. controls (385.33,179.17) and (376,169.83) .. (376,158.32) -- cycle ;
\draw   (376,208.96) .. controls (376,197.44) and (385.33,188.11) .. (396.85,188.11) .. controls (408.36,188.11) and (417.7,197.44) .. (417.7,208.96) .. controls (417.7,220.47) and (408.36,229.81) .. (396.85,229.81) .. controls (385.33,229.81) and (376,220.47) .. (376,208.96) -- cycle ;
\draw   (376,286.41) .. controls (376,274.89) and (385.33,265.55) .. (396.85,265.55) .. controls (408.36,265.55) and (417.7,274.89) .. (417.7,286.41) .. controls (417.7,297.92) and (408.36,307.26) .. (396.85,307.26) .. controls (385.33,307.26) and (376,297.92) .. (376,286.41) -- cycle ;
\draw [line width=2.25]  [dash pattern={on 2.53pt off 3.02pt}]  (397.59,235.77) -- (397.59,261.83) ;
\draw  [fill={rgb, 255:red, 245; green, 166; blue, 35 }  ,fill opacity=0.61 ] (375.25,336.3) .. controls (375.25,324.79) and (384.59,315.45) .. (396.1,315.45) .. controls (407.62,315.45) and (416.96,324.79) .. (416.96,336.3) .. controls (416.96,347.82) and (407.62,357.15) .. (396.1,357.15) .. controls (384.59,357.15) and (375.25,347.82) .. (375.25,336.3) -- cycle ;
\draw  [fill={rgb, 255:red, 74; green, 144; blue, 226 }  ,fill opacity=0.51 ] (472.81,158.32) .. controls (472.81,146.8) and (482.14,137.47) .. (493.66,137.47) .. controls (505.18,137.47) and (514.51,146.8) .. (514.51,158.32) .. controls (514.51,169.83) and (505.18,179.17) .. (493.66,179.17) .. controls (482.14,179.17) and (472.81,169.83) .. (472.81,158.32) -- cycle ;
\draw  [fill={rgb, 255:red, 74; green, 144; blue, 226 }  ,fill opacity=0.51 ] (472.81,208.96) .. controls (472.81,197.44) and (482.14,188.11) .. (493.66,188.11) .. controls (505.18,188.11) and (514.51,197.44) .. (514.51,208.96) .. controls (514.51,220.47) and (505.18,229.81) .. (493.66,229.81) .. controls (482.14,229.81) and (472.81,220.47) .. (472.81,208.96) -- cycle ;
\draw  [fill={rgb, 255:red, 74; green, 144; blue, 226 }  ,fill opacity=0.51 ] (472.81,286.41) .. controls (472.81,274.89) and (482.14,265.55) .. (493.66,265.55) .. controls (505.18,265.55) and (514.51,274.89) .. (514.51,286.41) .. controls (514.51,297.92) and (505.18,307.26) .. (493.66,307.26) .. controls (482.14,307.26) and (472.81,297.92) .. (472.81,286.41) -- cycle ;
\draw [line width=2.25]  [dash pattern={on 2.53pt off 3.02pt}]  (494.4,235.77) -- (494.4,261.83) ;
\draw  [fill={rgb, 255:red, 245; green, 166; blue, 35 }  ,fill opacity=0.61 ] (472.06,336.3) .. controls (472.06,324.79) and (481.4,315.45) .. (492.92,315.45) .. controls (504.43,315.45) and (513.77,324.79) .. (513.77,336.3) .. controls (513.77,347.82) and (504.43,357.15) .. (492.92,357.15) .. controls (481.4,357.15) and (472.06,347.82) .. (472.06,336.3) -- cycle ;
\draw    (417.7,158.32) -- (470.81,158.32) ;
\draw [shift={(472.81,158.32)}, rotate = 180] [color={rgb, 255:red, 0; green, 0; blue, 0 }  ][line width=0.75]    (10.93,-3.29) .. controls (6.95,-1.4) and (3.31,-0.3) .. (0,0) .. controls (3.31,0.3) and (6.95,1.4) .. (10.93,3.29)   ;
\draw    (417.7,158.32) -- (471.34,207.6) ;
\draw [shift={(472.81,208.96)}, rotate = 222.57999999999998] [color={rgb, 255:red, 0; green, 0; blue, 0 }  ][line width=0.75]    (10.93,-3.29) .. controls (6.95,-1.4) and (3.31,-0.3) .. (0,0) .. controls (3.31,0.3) and (6.95,1.4) .. (10.93,3.29)   ;
\draw    (417.7,158.32) -- (472.02,284.57) ;
\draw [shift={(472.81,286.41)}, rotate = 246.72] [color={rgb, 255:red, 0; green, 0; blue, 0 }  ][line width=0.75]    (10.93,-3.29) .. controls (6.95,-1.4) and (3.31,-0.3) .. (0,0) .. controls (3.31,0.3) and (6.95,1.4) .. (10.93,3.29)   ;
\draw [color={rgb, 255:red, 65; green, 117; blue, 5 }  ,draw opacity=1 ]   (417.7,208.96) -- (471.34,159.67) ;
\draw [shift={(472.81,158.32)}, rotate = 497.42] [color={rgb, 255:red, 65; green, 117; blue, 5 }  ,draw opacity=1 ][line width=0.75]    (10.93,-3.29) .. controls (6.95,-1.4) and (3.31,-0.3) .. (0,0) .. controls (3.31,0.3) and (6.95,1.4) .. (10.93,3.29)   ;
\draw [color={rgb, 255:red, 65; green, 117; blue, 5 }  ,draw opacity=1 ]   (417.7,208.96) -- (470.81,208.96) ;
\draw [shift={(472.81,208.96)}, rotate = 180] [color={rgb, 255:red, 65; green, 117; blue, 5 }  ,draw opacity=1 ][line width=0.75]    (10.93,-3.29) .. controls (6.95,-1.4) and (3.31,-0.3) .. (0,0) .. controls (3.31,0.3) and (6.95,1.4) .. (10.93,3.29)   ;
\draw [color={rgb, 255:red, 65; green, 117; blue, 5 }  ,draw opacity=1 ]   (417.7,208.96) -- (471.65,284.78) ;
\draw [shift={(472.81,286.41)}, rotate = 234.57] [color={rgb, 255:red, 65; green, 117; blue, 5 }  ,draw opacity=1 ][line width=0.75]    (10.93,-3.29) .. controls (6.95,-1.4) and (3.31,-0.3) .. (0,0) .. controls (3.31,0.3) and (6.95,1.4) .. (10.93,3.29)   ;
\draw [color={rgb, 255:red, 144; green, 19; blue, 254 }  ,draw opacity=1 ]   (417.7,286.41) -- (470.81,286.41) ;
\draw [shift={(472.81,286.41)}, rotate = 180] [color={rgb, 255:red, 144; green, 19; blue, 254 }  ,draw opacity=1 ][line width=0.75]    (10.93,-3.29) .. controls (6.95,-1.4) and (3.31,-0.3) .. (0,0) .. controls (3.31,0.3) and (6.95,1.4) .. (10.93,3.29)   ;
\draw [color={rgb, 255:red, 144; green, 19; blue, 254 }  ,draw opacity=1 ]   (417.7,286.41) -- (471.65,210.59) ;
\draw [shift={(472.81,208.96)}, rotate = 485.43] [color={rgb, 255:red, 144; green, 19; blue, 254 }  ,draw opacity=1 ][line width=0.75]    (10.93,-3.29) .. controls (6.95,-1.4) and (3.31,-0.3) .. (0,0) .. controls (3.31,0.3) and (6.95,1.4) .. (10.93,3.29)   ;
\draw [color={rgb, 255:red, 144; green, 19; blue, 254 }  ,draw opacity=1 ]   (417.7,286.41) -- (472.02,160.15) ;
\draw [shift={(472.81,158.32)}, rotate = 473.28] [color={rgb, 255:red, 144; green, 19; blue, 254 }  ,draw opacity=1 ][line width=0.75]    (10.93,-3.29) .. controls (6.95,-1.4) and (3.31,-0.3) .. (0,0) .. controls (3.31,0.3) and (6.95,1.4) .. (10.93,3.29)   ;
\draw [color={rgb, 255:red, 126; green, 211; blue, 33 }  ,draw opacity=1 ]   (416.96,336.3) -- (472.21,160.23) ;
\draw [shift={(472.81,158.32)}, rotate = 467.42] [color={rgb, 255:red, 126; green, 211; blue, 33 }  ,draw opacity=1 ][line width=0.75]    (10.93,-3.29) .. controls (6.95,-1.4) and (3.31,-0.3) .. (0,0) .. controls (3.31,0.3) and (6.95,1.4) .. (10.93,3.29)   ;
\draw [color={rgb, 255:red, 126; green, 211; blue, 33 }  ,draw opacity=1 ]   (416.96,336.3) -- (472,210.79) ;
\draw [shift={(472.81,208.96)}, rotate = 473.68] [color={rgb, 255:red, 126; green, 211; blue, 33 }  ,draw opacity=1 ][line width=0.75]    (10.93,-3.29) .. controls (6.95,-1.4) and (3.31,-0.3) .. (0,0) .. controls (3.31,0.3) and (6.95,1.4) .. (10.93,3.29)   ;
\draw [color={rgb, 255:red, 126; green, 211; blue, 33 }  ,draw opacity=1 ]   (416.96,336.3) -- (471.32,287.74) ;
\draw [shift={(472.81,286.41)}, rotate = 498.22] [color={rgb, 255:red, 126; green, 211; blue, 33 }  ,draw opacity=1 ][line width=0.75]    (10.93,-3.29) .. controls (6.95,-1.4) and (3.31,-0.3) .. (0,0) .. controls (3.31,0.3) and (6.95,1.4) .. (10.93,3.29)   ;
\draw  [fill={rgb, 255:red, 126; green, 211; blue, 33 }  ,fill opacity=0.61 ] (556.96,208.96) .. controls (556.96,197.44) and (566.29,188.11) .. (577.81,188.11) .. controls (589.33,188.11) and (598.66,197.44) .. (598.66,208.96) .. controls (598.66,220.47) and (589.33,229.81) .. (577.81,229.81) .. controls (566.29,229.81) and (556.96,220.47) .. (556.96,208.96) -- cycle ;
\draw  [fill={rgb, 255:red, 126; green, 211; blue, 33 }  ,fill opacity=0.61 ] (556.21,275.98) .. controls (556.21,264.46) and (565.55,255.13) .. (577.07,255.13) .. controls (588.58,255.13) and (597.92,264.46) .. (597.92,275.98) .. controls (597.92,287.5) and (588.58,296.83) .. (577.07,296.83) .. controls (565.55,296.83) and (556.21,287.5) .. (556.21,275.98) -- cycle ;
\draw    (514.51,158.32) -- (555.67,207.42) ;
\draw [shift={(556.96,208.96)}, rotate = 230.03] [color={rgb, 255:red, 0; green, 0; blue, 0 }  ][line width=0.75]    (10.93,-3.29) .. controls (6.95,-1.4) and (3.31,-0.3) .. (0,0) .. controls (3.31,0.3) and (6.95,1.4) .. (10.93,3.29)   ;
\draw [color={rgb, 255:red, 74; green, 144; blue, 226 }  ,draw opacity=1 ]   (514.51,208.96) -- (554.96,208.96) ;
\draw [shift={(556.96,208.96)}, rotate = 180] [color={rgb, 255:red, 74; green, 144; blue, 226 }  ,draw opacity=1 ][line width=0.75]    (10.93,-3.29) .. controls (6.95,-1.4) and (3.31,-0.3) .. (0,0) .. controls (3.31,0.3) and (6.95,1.4) .. (10.93,3.29)   ;
\draw    (514.51,158.32) -- (555.55,274.1) ;
\draw [shift={(556.21,275.98)}, rotate = 250.48000000000002] [color={rgb, 255:red, 0; green, 0; blue, 0 }  ][line width=0.75]    (10.93,-3.29) .. controls (6.95,-1.4) and (3.31,-0.3) .. (0,0) .. controls (3.31,0.3) and (6.95,1.4) .. (10.93,3.29)   ;
\draw [line width=2.25]  [dash pattern={on 2.53pt off 3.02pt}]  (577.07,233.53) -- (577.07,253.64) ;
\draw [color={rgb, 255:red, 74; green, 144; blue, 226 }  ,draw opacity=1 ]   (514.51,208.96) -- (555.16,274.28) ;
\draw [shift={(556.21,275.98)}, rotate = 238.11] [color={rgb, 255:red, 74; green, 144; blue, 226 }  ,draw opacity=1 ][line width=0.75]    (10.93,-3.29) .. controls (6.95,-1.4) and (3.31,-0.3) .. (0,0) .. controls (3.31,0.3) and (6.95,1.4) .. (10.93,3.29)   ;
\draw [color={rgb, 255:red, 144; green, 19; blue, 254 }  ,draw opacity=1 ]   (514.51,286.41) -- (556,210.71) ;
\draw [shift={(556.96,208.96)}, rotate = 478.73] [color={rgb, 255:red, 144; green, 19; blue, 254 }  ,draw opacity=1 ][line width=0.75]    (10.93,-3.29) .. controls (6.95,-1.4) and (3.31,-0.3) .. (0,0) .. controls (3.31,0.3) and (6.95,1.4) .. (10.93,3.29)   ;
\draw [color={rgb, 255:red, 144; green, 19; blue, 254 }  ,draw opacity=1 ]   (514.51,286.41) -- (554.27,276.47) ;
\draw [shift={(556.21,275.98)}, rotate = 525.96] [color={rgb, 255:red, 144; green, 19; blue, 254 }  ,draw opacity=1 ][line width=0.75]    (10.93,-3.29) .. controls (6.95,-1.4) and (3.31,-0.3) .. (0,0) .. controls (3.31,0.3) and (6.95,1.4) .. (10.93,3.29)   ;
\draw [color={rgb, 255:red, 126; green, 211; blue, 33 }  ,draw opacity=1 ]   (513.77,336.3) -- (556.32,210.85) ;
\draw [shift={(556.96,208.96)}, rotate = 468.74] [color={rgb, 255:red, 126; green, 211; blue, 33 }  ,draw opacity=1 ][line width=0.75]    (10.93,-3.29) .. controls (6.95,-1.4) and (3.31,-0.3) .. (0,0) .. controls (3.31,0.3) and (6.95,1.4) .. (10.93,3.29)   ;
\draw [color={rgb, 255:red, 126; green, 211; blue, 33 }  ,draw opacity=1 ]   (513.77,336.3) -- (555.06,277.62) ;
\draw [shift={(556.21,275.98)}, rotate = 485.13] [color={rgb, 255:red, 126; green, 211; blue, 33 }  ,draw opacity=1 ][line width=0.75]    (10.93,-3.29) .. controls (6.95,-1.4) and (3.31,-0.3) .. (0,0) .. controls (3.31,0.3) and (6.95,1.4) .. (10.93,3.29)   ;
\draw   (169,246.03) .. controls (169,240.92) and (173.14,236.78) .. (178.24,236.78) -- (252.25,236.78) .. controls (257.35,236.78) and (261.49,240.92) .. (261.49,246.03) -- (261.49,273.76) .. controls (261.49,278.86) and (257.35,283) .. (252.25,283) -- (178.24,283) .. controls (173.14,283) and (169,278.86) .. (169,273.76) -- cycle ;
\draw   (344.24,142.55) .. controls (344.24,108.01) and (372.24,80) .. (406.79,80) -- (594.45,80) .. controls (628.99,80) and (657,108.01) .. (657,142.55) -- (657,357.45) .. controls (657,391.99) and (628.99,420) .. (594.45,420) -- (406.79,420) .. controls (372.24,420) and (344.24,391.99) .. (344.24,357.45) -- cycle ;
\draw   (168,183.03) .. controls (168,177.92) and (172.14,173.78) .. (177.24,173.78) -- (251.25,173.78) .. controls (256.35,173.78) and (260.49,177.92) .. (260.49,183.03) -- (260.49,210.76) .. controls (260.49,215.86) and (256.35,220) .. (251.25,220) -- (177.24,220) .. controls (172.14,220) and (168,215.86) .. (168,210.76) -- cycle ;
\draw   (172,310.03) .. controls (172,304.92) and (176.14,300.78) .. (181.24,300.78) -- (255.25,300.78) .. controls (260.35,300.78) and (264.49,304.92) .. (264.49,310.03) -- (264.49,337.76) .. controls (264.49,342.86) and (260.35,347) .. (255.25,347) -- (181.24,347) .. controls (176.14,347) and (172,342.86) .. (172,337.76) -- cycle ;
\draw   (44,171.4) .. controls (44,145.22) and (65.22,124) .. (91.4,124) -- (233.6,124) .. controls (259.78,124) and (281,145.22) .. (281,171.4) -- (281,325.6) .. controls (281,351.78) and (259.78,373) .. (233.6,373) -- (91.4,373) .. controls (65.22,373) and (44,351.78) .. (44,325.6) -- cycle ;
\draw [line width=1.25]    (-6,277) -- (39,277) ;
\draw [shift={(43,277)}, rotate = 180] [color={rgb, 255:red, 0; green, 0; blue, 0 }  ][line width=1.25]    (17.49,-5.26) .. controls (11.12,-2.23) and (5.29,-0.48) .. (0,0) .. controls (5.29,0.48) and (11.12,2.23) .. (17.49,5.26)   ;
\draw [color={rgb, 255:red, 208; green, 2; blue, 27 }  ,draw opacity=1 ][fill={rgb, 255:red, 208; green, 2; blue, 27 }  ,fill opacity=1 ]   (43,277) -- (166.23,211.69) ;
\draw [shift={(168,210.76)}, rotate = 512.0799999999999] [color={rgb, 255:red, 208; green, 2; blue, 27 }  ,draw opacity=1 ][line width=0.75]    (10.93,-3.29) .. controls (6.95,-1.4) and (3.31,-0.3) .. (0,0) .. controls (3.31,0.3) and (6.95,1.4) .. (10.93,3.29)   ;
\draw [color={rgb, 255:red, 74; green, 144; blue, 226 }  ,draw opacity=1 ]   (43,277) -- (167.51,262.24) ;
\draw [shift={(169.5,262)}, rotate = 533.24] [color={rgb, 255:red, 74; green, 144; blue, 226 }  ,draw opacity=1 ][line width=0.75]    (10.93,-3.29) .. controls (6.95,-1.4) and (3.31,-0.3) .. (0,0) .. controls (3.31,0.3) and (6.95,1.4) .. (10.93,3.29)   ;
\draw [color={rgb, 255:red, 189; green, 16; blue, 224 }  ,draw opacity=1 ]   (43,277) -- (171.63,325.3) ;
\draw [shift={(173.5,326)}, rotate = 200.57999999999998] [color={rgb, 255:red, 189; green, 16; blue, 224 }  ,draw opacity=1 ][line width=0.75]    (10.93,-3.29) .. controls (6.95,-1.4) and (3.31,-0.3) .. (0,0) .. controls (3.31,0.3) and (6.95,1.4) .. (10.93,3.29)   ;
\draw [color={rgb, 255:red, 208; green, 2; blue, 27 }  ,draw opacity=1 ]   (261.5,200) -- (374.12,159) ;
\draw [shift={(376,158.32)}, rotate = 520] [color={rgb, 255:red, 208; green, 2; blue, 27 }  ,draw opacity=1 ][line width=0.75]    (10.93,-3.29) .. controls (6.95,-1.4) and (3.31,-0.3) .. (0,0) .. controls (3.31,0.3) and (6.95,1.4) .. (10.93,3.29)   ;
\draw [color={rgb, 255:red, 74; green, 144; blue, 226 }  ,draw opacity=1 ]   (261.5,259) -- (374.49,159.64) ;
\draw [shift={(376,158.32)}, rotate = 498.67] [color={rgb, 255:red, 74; green, 144; blue, 226 }  ,draw opacity=1 ][line width=0.75]    (10.93,-3.29) .. controls (6.95,-1.4) and (3.31,-0.3) .. (0,0) .. controls (3.31,0.3) and (6.95,1.4) .. (10.93,3.29)   ;
\draw [color={rgb, 255:red, 189; green, 16; blue, 224 }  ,draw opacity=1 ]   (265.5,323) -- (374.88,159.98) ;
\draw [shift={(376,158.32)}, rotate = 483.86] [color={rgb, 255:red, 189; green, 16; blue, 224 }  ,draw opacity=1 ][line width=0.75]    (10.93,-3.29) .. controls (6.95,-1.4) and (3.31,-0.3) .. (0,0) .. controls (3.31,0.3) and (6.95,1.4) .. (10.93,3.29)   ;
\draw [color={rgb, 255:red, 208; green, 2; blue, 27 }  ,draw opacity=1 ]   (261.5,200) -- (374,208.8) ;
\draw [shift={(376,208.96)}, rotate = 184.47] [color={rgb, 255:red, 208; green, 2; blue, 27 }  ,draw opacity=1 ][line width=0.75]    (10.93,-3.29) .. controls (6.95,-1.4) and (3.31,-0.3) .. (0,0) .. controls (3.31,0.3) and (6.95,1.4) .. (10.93,3.29)   ;
\draw [color={rgb, 255:red, 208; green, 2; blue, 27 }  ,draw opacity=1 ]   (261.5,200) -- (374.4,285.2) ;
\draw [shift={(376,286.41)}, rotate = 217.04] [color={rgb, 255:red, 208; green, 2; blue, 27 }  ,draw opacity=1 ][line width=0.75]    (10.93,-3.29) .. controls (6.95,-1.4) and (3.31,-0.3) .. (0,0) .. controls (3.31,0.3) and (6.95,1.4) .. (10.93,3.29)   ;
\draw [color={rgb, 255:red, 74; green, 144; blue, 226 }  ,draw opacity=1 ]   (261.5,259) -- (295.08,244.32) -- (374.16,209.76) ;
\draw [shift={(376,208.96)}, rotate = 516.39] [color={rgb, 255:red, 74; green, 144; blue, 226 }  ,draw opacity=1 ][line width=0.75]    (10.93,-3.29) .. controls (6.95,-1.4) and (3.31,-0.3) .. (0,0) .. controls (3.31,0.3) and (6.95,1.4) .. (10.93,3.29)   ;
\draw [color={rgb, 255:red, 74; green, 144; blue, 226 }  ,draw opacity=1 ]   (261.5,259) -- (374.05,285.94) ;
\draw [shift={(376,286.41)}, rotate = 193.46] [color={rgb, 255:red, 74; green, 144; blue, 226 }  ,draw opacity=1 ][line width=0.75]    (10.93,-3.29) .. controls (6.95,-1.4) and (3.31,-0.3) .. (0,0) .. controls (3.31,0.3) and (6.95,1.4) .. (10.93,3.29)   ;
\draw [color={rgb, 255:red, 189; green, 16; blue, 224 }  ,draw opacity=1 ]   (265.5,323) -- (374.61,210.39) ;
\draw [shift={(376,208.96)}, rotate = 494.1] [color={rgb, 255:red, 189; green, 16; blue, 224 }  ,draw opacity=1 ][line width=0.75]    (10.93,-3.29) .. controls (6.95,-1.4) and (3.31,-0.3) .. (0,0) .. controls (3.31,0.3) and (6.95,1.4) .. (10.93,3.29)   ;
\draw [color={rgb, 255:red, 189; green, 16; blue, 224 }  ,draw opacity=1 ]   (265.5,323) -- (374.1,287.04) ;
\draw [shift={(376,286.41)}, rotate = 521.6800000000001] [color={rgb, 255:red, 189; green, 16; blue, 224 }  ,draw opacity=1 ][line width=0.75]    (10.93,-3.29) .. controls (6.95,-1.4) and (3.31,-0.3) .. (0,0) .. controls (3.31,0.3) and (6.95,1.4) .. (10.93,3.29)   ;

\draw (387.95,147.32) node [anchor=north west][inner sep=0.75pt]   [align=left] {{\large \textbf{x'\textsubscript{1}}}};
\draw (387.95,197.96) node [anchor=north west][inner sep=0.75pt]   [align=left] {{\large \textbf{x'\textsubscript{2}}}};
\draw (387.95,275.41) node [anchor=north west][inner sep=0.75pt]   [align=left] {{\large \textbf{x'\textsubscript{n}}}};
\draw (390.98,327.43) node [anchor=north west][inner sep=0.75pt]   [align=left] {\textbf{1}};
\draw (489.28,147.45) node [anchor=north west][inner sep=0.75pt]   [align=left] {\textbf{{\large f}}};
\draw (487.79,327.43) node [anchor=north west][inner sep=0.75pt]   [align=left] {\textbf{1}};
\draw (488.53,199.57) node [anchor=north west][inner sep=0.75pt]   [align=left] {\textbf{{\large f}}};
\draw (488.53,276.28) node [anchor=north west][inner sep=0.75pt]   [align=left] {\textbf{{\large f}}};
\draw (423.02,124.13) node [anchor=north west][inner sep=0.75pt]   [align=left] {\textbf{W\textsuperscript{1},b\textsuperscript{1}}};
\draw (570.94,198.09) node [anchor=north west][inner sep=0.75pt]   [align=left] {\textbf{{\large g}}};
\draw (570.19,265.11) node [anchor=north west][inner sep=0.75pt]   [align=left] {\textbf{{\large g}}};
\draw (526.53,150.93) node [anchor=north west][inner sep=0.75pt]   [align=left] {\textbf{W\textsuperscript{2},b\textsuperscript{2}}};
\draw (597.73,232.96) node [anchor=north west][inner sep=0.75pt]   [align=left] {\textbf{{\large h\textsubscript{w,b}(x)}}};
\draw (358.59,359) node [anchor=north west][inner sep=0.75pt]   [align=left] {\begin{minipage}[lt]{54.304392pt}\setlength\topsep{0pt}
	\begin{center}
	\textbf{Input layer}\\\textbf{(n units)}
	\end{center}
	
	\end{minipage}};
\draw (455.85,359) node [anchor=north west][inner sep=0.75pt]   [align=left] {\begin{minipage}[lt]{63.941284pt}\setlength\topsep{0pt}
	\begin{center}
	\textbf{Hidden layer}\\\textbf{(m neurons)}
	\end{center}
	
	\end{minipage}};
\draw (532.07,298.68) node [anchor=north west][inner sep=0.75pt]   [align=left] {\begin{minipage}[lt]{62.804392pt}\setlength\topsep{0pt}
	\begin{center}
	\textbf{Output layer}\\\textbf{(r outputs)}
	\end{center}
	
	\end{minipage}};
\draw (169.41,177.97) node [anchor=north west][inner sep=0.75pt]   [align=left] {\begin{minipage}[lt]{63.941284pt}\setlength\topsep{0pt}
	\begin{center}
	\textbf{{\small Differentiable }}\\\textbf{{\small transformer}}
	\end{center}
	
	\end{minipage}};
\draw (171.31,239.78) node [anchor=north west][inner sep=0.75pt]   [align=left] {\begin{minipage}[lt]{63.941284pt}\setlength\topsep{0pt}
	\begin{center}
	\textbf{{\small Differentiable }}\\\textbf{{\small grid}}
	\end{center}
	
	\end{minipage}};
\draw (438.09,91.2) node [anchor=north west][inner sep=0.75pt]   [align=left] {\begin{minipage}[lt]{85.46784000000001pt}\setlength\topsep{0pt}
	\begin{center}
	\textbf{{\large Fixed Weights}}
	\end{center}
	
	\end{minipage}};
\draw (2.95,241.32) node [anchor=north west][inner sep=0.75pt]   [align=left] {\textbf{{\LARGE X}}};
\draw (191.31,304.78) node [anchor=north west][inner sep=0.75pt]   [align=left] {\begin{minipage}[lt]{39.004392pt}\setlength\topsep{0pt}
	\begin{center}
	\textbf{Smart }\\\textbf{erasing}
	\end{center}
	
	\end{minipage}};
\draw (63.09,139.2) node [anchor=north west][inner sep=0.75pt]   [align=left] {\begin{minipage}[lt]{134.66108000000003pt}\setlength\topsep{0pt}
	\begin{center}
	\textbf{{\large Trainable Components}}
	\end{center}
	
	\end{minipage}};
\draw (52,178) node [anchor=north west][inner sep=0.75pt]   [align=left] {\begin{minipage}[lt]{51.4675pt}\setlength\topsep{0pt}
	\begin{center}
	\textbf{randomly }\\\textbf{select a }\\\textbf{strategy}
	\end{center}
	
	\end{minipage}};
\draw (300.95,156.32) node [anchor=north west][inner sep=0.75pt]   [align=left] {\textbf{{\LARGE X'}}};

\end{tikzpicture}

	\caption{Adversarial data synthesizer network}
	\label{adv_gen}
\end{figure}
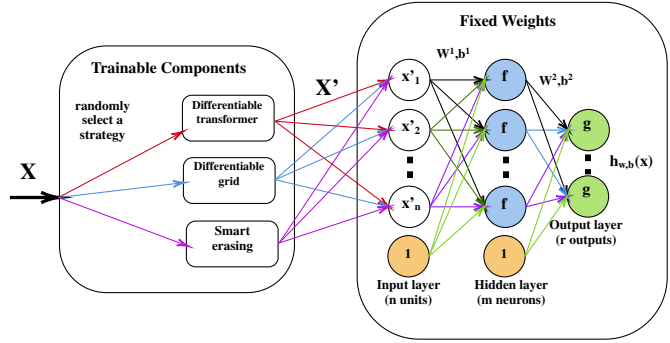

	The main idea of SCNN is to fix the weights of the embedded neural network, and use the gradient descent technique in order to manipulate the input data and generate a set of adversarial data. As an example, the architecture of the ADS network with MLP as the embedded network is presented in Fig.~\ref{adv_gen}. In this structure, the input vector $X = [x_{1},x_{2},...,x_{n}]$ is fed to a (a set of) differentiable component(s) that try to manipulate the input data to a vector $X' = [x'_{1},x'_{2},...,x'_{n}]$, which is fed to the embedded network. To do that, we define an optimization problem, where the parameters of the differentiable components are trained based on a cost function that is defined in a way that the manipulated data is predicted as the nearest wrong label by the embedded neural network.

	The trainable components are designed based on the type of the input dataset. In this paper, we consider three main trainable components for the {\em image} samples: 

	\begin{itemize}
		\item \textbf{Differentiable transformer:} We use Spatial Transformer Network (STN) layer  introduced in~\cite{jaderberg2015spatial} for this transformation. It takes a vector of size 6 per image to perform an affine transformation (such as rotation, zoom, shearing) on the input image. For example, if we have an input image and the coordinate of each pixel is identified by $ (x_{i}^{s},y_{i}^{s}) $, and $ (x_{i}^{t},y_{i}^{t})$ represents the coordinate of the pixel after the affine transformation, we have:
		
		\begin{equation}
		\label{eq8}
		G_t = \tau_{\theta}(G_{i})=
		\begin{pmatrix}
			x_{i}^{t} \\
			y_{i}^{t}
		\end{pmatrix}
		=
		\begin{bmatrix}
		\theta_{11} & \theta_{12} & \theta_{13}  \\
		\theta_{21} & \theta_{22} & \theta_{23}
		\end{bmatrix}
		\begin{pmatrix}
		x_{i}^{s} \\
		y_{i}^{s} \\
		1
		\end{pmatrix}
		\end{equation}

%
%

		, where $G_{i}$ is the initial grid of the input feature map, and  $G_{t}$ is the final grid of the feature map after applying the transformation $\tau_{\theta}$ on $G_{i}$. 
		The most important feature of this component is that it is differentiable with respect to its transformation parameters, which means the gradient descent can learn them based on a defined cost function.  
		\item \textbf{Differentiable grid:} We design this component to be more flexible to perform grid manipulation in the input feature map. It is a differentiable component that manipulates the grid of the input image in three main steps:
		\begin{enumerate}
			\item Convolution on a null input space: For each input image, we generate a null input space ($\delta$) based on sampling from uniform distribution $U(-1,+1)$ with the same size as the input image size, denoted by $(W,H)$ . This space is the starting point for grid manipulation, and  will be trained during the adversarial data synthesis phase. After that, we define a Gaussian kernel window, which is convolved with this null space, and as a result, we will have a smoothed null input space, as depicted in Fig.~\ref{diffgrid1}. 
			The convolved null space ($\mu$) can be computed as follows:
			\begin{equation}
			\label{eq15}
			\begin{split}
			& \delta_{ij} \sim U(-1,+1) : \; i \in [1...W] \; \; j \in [1...H] \\ 
			& \mu_{ij}= \delta_{ij} \circledast Kernel_{ij} =  \sum\limits_{m=1}^{w_K} \sum\limits_{n=1}^{h_k} Kernel_{m,n}(\Phi_{\mu},\Phi_{\sigma})\delta_{i-m,j-n}:\\ 
			&\forall i \in [1...W] \; \; \forall j \in [1...H] 
			\end{split}
			\end{equation}
	, where $Kernel$ is a predefined 2d Gaussian kernel with size $(w_K,h_K)$, and mean and standard deviation equal to $\Phi_{\mu}$ and $\Phi_{\sigma}$, respectively. Note that this kernel is immutable during the training phase of the input null space.

			\item Generating the manipulated grid:  In this step, the convolved null space $(\mu)$ is added to the initial grid $(G_{i})$ of the input image, and as a result, the manipulated grid $(G_{t})$ is generated, as shown in Fig.~\ref{diffgrid2}. 
			
						\begin{equation}
			\label{eq16}
			\begin{split}
			& G_{t}=G_{i}+\mu
			\end{split}
			\end{equation}

		 	\item Image sampling: The main operation of this step is an image interpolation based on the manipulated grid generated in the previous step. As depicted in Fig.~\ref{diffgrid3}, the image sampler takes an image and a manipulated grid as input, and computes the interpolated output image based on the manipulated grid.  The Final output image can be computed as follows: 
		 	
		 	\begin{equation}
		 	\label{eq17}
		 	\begin{split}
		 	&O_{i}^{c}=
		 	\sum\limits_{n=1}^H \sum\limits_{m=1}^W I_{nm}^{c}k(x_{i}^{t}-m;\Phi_{x})k(y_{i}^{t}-n;\Phi_{y})\\     
		 	&\forall i \in [1...HW] \; \forall c \in [1...C]\\
		 	&(x^{t},y^{t}) \in G_{t}
		 	\end{split}
		 	\end{equation}

		 	, where $k$ is a generic sampling kernel function that performs the image interpolation. It can be any image interpolation function (e.g bilinear), and $\Phi_{x}$ and $\Phi_{y}$ are the parameters of this kernel function. $I_{nm}^{c}$ is the input value at location $(n,m)$ in channel $c$ of the input image, and $O_{i}^{c}$ is the output value for pixel $i$ at location $(x_{i}^{t},y_{i}^{t})$. For example, we can use a bilinear sampling kernel, as follows: 
		 	
		 	\begin{equation}
		 	\label{eq10}
		 	\begin{split}
		 	&O_{i}^{c}=\sum\limits_{n=1}^H \sum\limits_{m=1}^W I_{nm}^{c}max(0,1-|x_{i}^{t} - m|)max (0,1-|y_{i}^{t}-n|) \\ 
		 	&\forall i \in [1...HW] \; \forall c \in [1...C]
		 	\end{split}
		 	\end{equation} 
		\end{enumerate}

		It can be proved that this component is differentiable with respect to its input grid $G^{t}$:

		\begin{equation}
		\label{eq11}
		\begin{split}
		&\frac{\partial O_{i}^{c}}{\partial x_{i}^{t}} = \sum\limits_{n=1}^H \sum\limits_{m=1}^W I_{nm}^{c}max (0,1-|y_{i}^{t}-n|)
		\left\{ \begin{array}{rcl}
		0 & if \;\; |x_{i}^{t} - m| \geq 1 \\ 
		1 &  if \;\; m \geq x_{i}^{t} \\
		-1 & if \;\; m < x_{i}^{t}
		\end{array}\right.
		\end{split}
		\end{equation} 
		
		Eq.~\ref{eq11} can be defined similarly for $\frac{\partial O_{i}^{c}}{\partial y_{i}^{t}}$.

		\begin{figure*}[h!]
	\centering
	\begin{subfigure}[t]{0.4\textwidth}
		\centering
		\begin{tikzpicture}[scale=0.6, every node/.style={scale=0.6},x=0.75pt,y=0.75pt,yscale=-1,xscale=1]
		
		\draw (72.15,217.37) node [rotate=-89.85,xslant=-0.56] {\includegraphics[width=77.18pt,height=62.93pt]{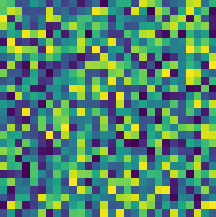}};
		\draw  [draw opacity=0][fill={rgb, 255:red, 255; green, 255; blue, 255 }  ,fill opacity=0.68 ] (151.77,224.23) -- (152.26,260.31) -- (123.69,276.05) -- (123.2,239.97) -- cycle ; \draw   (151.77,224.23) -- (123.2,239.97)(151.87,231.26) -- (123.3,247)(151.96,238.3) -- (123.4,254.04)(152.06,245.33) -- (123.49,261.08)(152.15,252.37) -- (123.59,268.11)(152.25,259.4) -- (123.68,275.15) ; \draw   (151.77,224.23) -- (152.26,260.31)(144.79,228.07) -- (145.28,264.16)(137.8,231.92) -- (138.29,268.01)(130.82,235.77) -- (131.31,271.86)(123.83,239.62) -- (124.32,275.71) ; \draw    ;
		\draw (224.28,263.09) node [rotate=-90.02,xslant=-0.55] {\includegraphics[width=77.31pt,height=62.83pt]{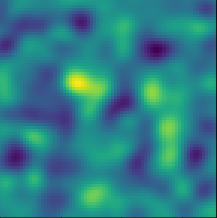}};
		\draw  [fill={rgb, 255:red, 255; green, 255; blue, 255 }  ,fill opacity=0.48 ] (113.91,142.3) -- (151.77,224.23) -- (123.83,239.62) -- (30.12,189.53) -- cycle ;
		\draw  [fill={rgb, 255:red, 255; green, 255; blue, 255 }  ,fill opacity=0.48 ] (30.12,189.53) -- (123.83,239.62) -- (124.31,274.8) -- (30.39,292.44) -- cycle ;
		\draw  [fill={rgb, 255:red, 179; green, 130; blue, 130 }  ,fill opacity=0.48 ] (151.77,224.23) -- (266.2,188.71) -- (182.41,234.4) -- (123.63,239) -- cycle ;
		\draw  [fill={rgb, 255:red, 133; green, 93; blue, 93 }  ,fill opacity=0.31 ] (124.31,239.62) -- (182.89,234.39) -- (182.85,337.47) -- (124.31,274.8) -- cycle ;
		
		\draw (144.01,201.2) node [anchor=north west][inner sep=0.75pt]  [rotate=-330.55] [align=left] {{\small Gaussian Kernel}};
		\draw (7.33,179.73) node [anchor=north west][inner sep=0.75pt]  [rotate=-330.55] [align=left] {Null input space};
		\draw (170.38,350.12) node [anchor=north west][inner sep=0.75pt]  [rotate=-331.08] [align=left] {Convolved null input space};
		\draw (75.28,291.2) node [anchor=north west][inner sep=0.75pt]  [rotate=-17.01] [align=left] {2d convolution};

		\end{tikzpicture}
		\caption{First Step of grid manipulation: Null input-space convolved with  predefined Gaussian kernel} 
		\label{diffgrid1}
	\end{subfigure} \hspace{0.1\textwidth} %
	~ 
	\begin{subfigure}[t]{0.4\textwidth}
		\centering
		\begin{tikzpicture}[scale=0.6, every node/.style={scale=0.6},x=0.75pt,y=0.75pt,yscale=-1,xscale=1]
		
		\draw  [draw opacity=0] (252.18,447.37) -- (252.25,530.84) -- (177.41,574.12) -- (177.34,490.65) -- cycle ; \draw   (252.18,447.37) -- (177.34,490.65)(252.19,455.68) -- (177.35,498.96)(252.2,464) -- (177.35,507.28)(252.2,472.32) -- (177.36,515.6)(252.21,480.63) -- (177.37,523.91)(252.22,488.95) -- (177.37,532.23)(252.23,497.27) -- (177.38,540.55)(252.23,505.58) -- (177.39,548.86)(252.24,513.9) -- (177.4,557.18)(252.25,522.22) -- (177.4,565.5)(252.25,530.53) -- (177.41,573.81) ; \draw   (252.18,447.37) -- (252.25,530.84)(243.87,452.17) -- (243.94,535.64)(235.56,456.98) -- (235.63,540.45)(227.25,461.79) -- (227.32,545.26)(218.93,466.59) -- (219.01,550.07)(210.62,471.4) -- (210.69,554.87)(202.31,476.21) -- (202.38,559.68)(194,481.02) -- (194.07,564.49)(185.68,485.82) -- (185.76,569.29)(177.37,490.63) -- (177.44,574.1) ; \draw    ;
		\draw (141.39,484.04) node [rotate=-90.02,xslant=-0.55] {\includegraphics[width=64.29pt,height=54.01pt]{convolved.png}};
		\draw (294.23,532.48) node [rotate=-90,xslant=-0.54] {\includegraphics[width=65.64pt,height=59.27pt]{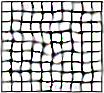}};
		
		\draw (148.68,536.89) node [anchor=north west][inner sep=0.75pt]  [font=\LARGE,rotate=0,xslant=0] [align=left] {\textbf{{\Large +}}};
		\draw (237.18,561.29) node [anchor=north west][inner sep=0.75pt]  [font=\LARGE,rotate=-19.91,xslant=0.79] [align=left] {\textbf{{\LARGE =}}};
		\draw (175.25,464.87) node [anchor=north west][inner sep=0.75pt]  [rotate=-331.08] [align=left] {Initial input grid};
		\draw (74.41,432.57) node [anchor=north west][inner sep=0.75pt]  [rotate=-331.08] [align=left] {Convolved\\  null input space};
		\draw (251.86,488.56) node [anchor=north west][inner sep=0.75pt]  [rotate=-331.08] [align=left] {manipulated grid};

		\end{tikzpicture}
		\caption{Second Step of grid manipulation: summation of convolved input space and initial grid of input image} 
		\label{diffgrid2}
		
	\end{subfigure} \par\bigskip 
	~ 
	\begin{subfigure}[t]{0.5\textwidth}
		\centering
		\begin{tikzpicture}[scale=0.6, every node/.style={scale=0.6},x=0.75pt,y=0.75pt,yscale=-1,xscale=1]
		
		\draw (314.05,447) node  {\includegraphics[width=68.92pt,height=61.5pt]{final-grid.png}};
		\draw (125.48,445.73) node [rotate=-90] {\includegraphics[width=60.76pt,height=67.58pt]{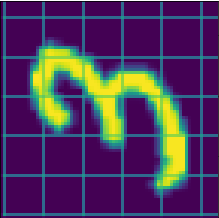}};
		\draw  [line width=1.25]  (206.31,446.5) .. controls (206.31,440.01) and (212.13,434.74) .. (219.32,434.74) .. controls (226.5,434.74) and (232.33,440.01) .. (232.33,446.5) .. controls (232.33,453) and (226.5,458.27) .. (219.32,458.27) .. controls (212.13,458.27) and (206.31,453) .. (206.31,446.5) -- cycle ;
		\draw [line width=1.25]    (219.32,434.74) -- (219.32,458.27) ;
		\draw [line width=1.25]    (206.31,446.5) -- (232.33,446.5) ;
		
		\draw (461.81,447.5) node [rotate=-90] {\includegraphics[width=60.75pt,height=68.72pt]{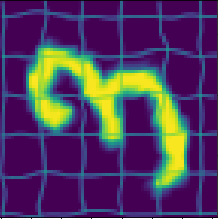}};
		
		\draw (381.69,448.57) node [anchor=north west][inner sep=0.75pt]  [font=\LARGE] [align=left] {\textbf{{\LARGE =}}};
		\draw (91.2,382.9) node [anchor=north west][inner sep=0.75pt]   [align=left] {input image};
		\draw (194.88,389.66) node [anchor=north west][inner sep=0.75pt]   [align=left] {\begin{minipage}[lt]{35.370608000000004pt}\setlength\topsep{0pt}
			\begin{center}
			{\small image }\\{\small sampler}
			\end{center}
			
			\end{minipage}};
		\draw (258.93,386.06) node [anchor=north west][inner sep=0.75pt]   [align=left] {\begin{minipage}[lt]{78.13608pt}\setlength\topsep{0pt}
			\begin{center}
			manipulated grid
			\end{center}
			
			\end{minipage}};
		\draw (420.8,387.54) node [anchor=north west][inner sep=0.75pt]   [align=left] {output image};

		\end{tikzpicture}
		\caption{Third Step of grid manipulation: Image sampling} 
		\label{diffgrid3}
	\end{subfigure}
	\caption{Three main steps of the differentiable grid component}
\end{figure*}

	\item \textbf{Smart erasing:} Random erasing is one of the most effective data augmentation techniques which is mostly used in DNNs \cite{OGara2019}. This component exploits erasing technique in order to smartly erase some parts of the input data in a way that the erased samples maximize the loss function. This component has four main steps: 
	 
		\begin{enumerate}
		\item Defining an NxN trainable mask $(M)$: this mask covers the input data and responsible for distinguishing which parts of the input have the most significant effect on classifying the input data. Parameter N could be varied based on input size(for example, for a 32x32 input size, N could be 4 or 8). 
		
		\item A 3d-Upsampling operation: this operation fits the size of the grid to the size of the input image. for instance, for a 4x4 grid, we use an 8x8x3 Upsampling operation in order to make a 32x32x3 grid(same size as the input data). The elements of the upsampled grid are multiplied to input data.

		\item Finding attention area: From the previous step, we prepare a trainable grid that covers the input data.  Now, we train this grid in order to distinguish which parts of the input data have the most significant effect on classifying tasks (Fig.~\ref{smart_erasing}.b).

		\item Erasing the most effective grids: after finding attention areas, grids with the most impact on the classifying task are deleted or replaced with noisy data.(Fig.~\ref{smart_erasing}.d, Fig.~\ref{smart_erasing}.e). There are two possibilities, if the removed area is a part of the background, it will help classifying task that decreases the importance of this part. If the removed area is an essential part of the input, the algorithm tries to identify the input without considering this part. 
		\end{enumerate}
	
	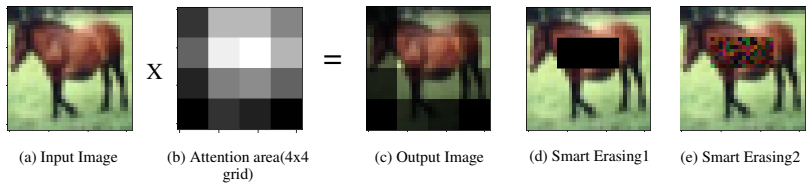
\begin{figure*}[h!]
	\centering
	\begin{tikzpicture}[scale=0.6, every node/.style={scale=0.6},x=0.75pt,y=0.75pt,yscale=-1,xscale=1]
	
	\draw (67.5,133.5) node  {\includegraphics[width=80.25pt,height=80.25pt]{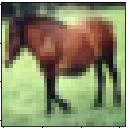}};
	\draw (210,135) node  {\includegraphics[width=79.5pt,height=79.5pt]{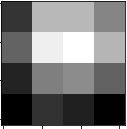}};
	\draw (368.5,134.5) node  {\includegraphics[width=78.75pt,height=78.75pt]{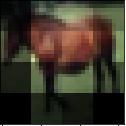}};
	\draw (503.5,134.5) node  {\includegraphics[width=78.75pt,height=78.75pt]{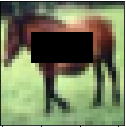}};
	\draw (632.5,134.5) node  {\includegraphics[width=78.75pt,height=78.75pt]{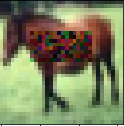}};
	
	\draw (23,202) node [anchor=north west][inner sep=0.75pt]   [align=left] {(a) Input Image};
	\draw (143,202) node [anchor=north west][inner sep=0.75pt]   [align=left] {\begin{minipage}[lt]{96.27304000000001pt}\setlength\topsep{0pt}
		\begin{center}
		(b) Attention area(4x4 grid)
		\end{center}
		
		\end{minipage}};
	\draw (130,129) node [anchor=north west][inner sep=0.75pt]   [align=left] {{\Large X}};
	\draw (279,121) node [anchor=north west][inner sep=0.75pt]   [align=left] {{\huge =}};
	\draw (322,202) node [anchor=north west][inner sep=0.75pt]   [align=left] {(c) Output Image};
	\draw (452,201) node [anchor=north west][inner sep=0.75pt]   [align=left] {(d) Smart Erasing1};
	\draw (579,202) node [anchor=north west][inner sep=0.75pt]   [align=left] {(e) Smart Erasing2};

	\end{tikzpicture}
		\caption{Applying smart erasing on a specific image of cifar10 database}
		\label{smart_erasing}
	\end{figure*}

	\end{itemize}
	We defined three components, the transformer component which is differentiable with respect to its transformation parameters $\theta$, and the grid manipulation component, which is differentiable with respect to its null input space $\delta$ and the smart erasing component which is differentiable with respect to its defined mask $(M)$. 
	Let's denote the output of the transformer component with $ V_{(X_j|\theta)}$ and the output of the grid manipulation component with $O_{(X_j|\delta)}$ and the output of the random erasing component with $E_{(X_j|M)}$  for input sample $ X_j $. 
	 $X'$ (the generated adversarial data to be fed to the embedded network) can be computed as follows (Fig.~\ref{adv_gen}):
		\begin{equation}
	\label{eq12}
	\begin{split}
	&X' =\left\{ \begin{array}{rcl}
	&V_{(X|\theta)} & or \\ 
	&O_{(X|\delta)} &  or \\
	&E_{(X|M)} &  
	\end{array}\right.
	\end{split}
	\end{equation}
	The corresponding function of the adversarial data synthesizer network is as follows:
	\begin{equation}
	\label{eq12a}
	\begin{split}
	&h^{ADS}(X|\theta,\delta) = h^{MLP}(X')
	\end{split}
	\end{equation}
	
	, where $h^{ADS}$ is the corresponding function of the ADS network, and  $h^{MLP}$ is the corresponding function of the embedded network (MLP in this example).
	The goal is to optimize parameters $\theta$, $\delta$ and $M$ in a way that the output of the ADS becomes the nearest wrong label for each input data. For defining an appropriate loss function, we need to define the following variables:
	
	\begin{equation}
	\label{eq13}
	\begin{split}
	&\hat{y}_{j}=\operatorname*{arg\,max}_{i} h_i(X_{j}|\theta,\delta)^{ADS}\\
	&\check{y}_{j}=\operatorname*{arg\,max}_{i} h_i(X_{j}|\theta,\delta)^{ADS}:i \neq \hat{y}_{j}\\
	&\bar{y}_{j}=\left\{ \begin{array}{rcl}
	&(y_{j}+ \alpha \check{y}_{j}) & if \; \hat{y}_{j} = y_j \\ 
	&(y_{j}+ \alpha \hat{y}_{j}) & if \; \hat{y}_{j} \neq y_j 
	\end{array}\right.
	\end{split}
	\end{equation}
	  
	, where $\hat{y}_{j}$ is the index of the largest output of the ADS network, $\check{y}_{j}$ is the index of the second largest output of the ADS network, and $\bar{y}_{j}$ is the desired output of the ADS network for generating harder samples. Note that $\bar{y}_{j}$ is constructed by a linear combination of the correct label $y_{j}$, and the nearest wrong label for the input data $X_{j}$. Also, $\alpha$ is the strength rate of moving the input data to the boundary of its true label and the nearest wrong label. Now, we can define the loss function of the ADS, as follows:
	\begin{equation}
	\label{eq14}
	loss^{ADS}=-\sum\limits_{i=1}^n \sum\limits_{k=1}^K \bar{y}_{ik}log(\hat{y}_{ik}) 
	\end{equation}
	
	The goal is to optimize parameters $(\theta,\delta)$, such that the manipulated input data is predicted with a wrong label, or becomes closer to the boundary of its true label and its nearest wrong label.

%% file: ExperimentalResults.tex
\section{Experimental Results}
\label{section:results}
We implemented SCNN on three datasets; MNIST, Fashion MNIST, and Cifar10.
The underlying embedded network is a CNN for MNIST and Fashion MNIST datasets, consisting of a convolutional, a BatchNormalization, and a pooling layer, repeated three times in a stack. Size of all filters is (3x3) and number of filters in convolution layers are 64,128,256 respectively. We also add a fully-connected hidden layer(with 512 neurons) to the end of last convolution layer. All activation functions are ReLu to have less intensive computation overhead in the training phase. For Cifar10 dataset we choose Resnet18 structure as our underlying embedded network. Number of filters in resnet18 start with 64 and we use a dropout layer(mask ratio:0.2) after the fully connected layer of Resnet18.

\subsection{Limited MNIST}
\begin{figure*}[htbp]
	\centerline{\includegraphics[width=\textwidth]{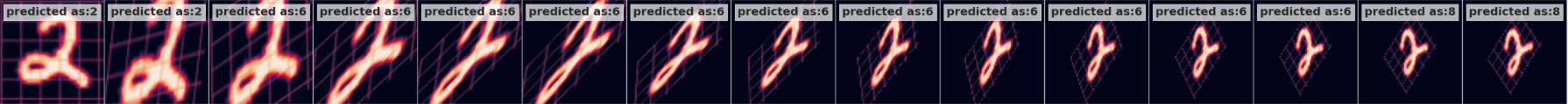}}
	\caption{Data manipulation based on differentiable transformer on MNIST sample data}
	\label{fig2}
\end{figure*}

We randomly selected 1000 samples of the MNIST handwritten dataset to have a more challenging problem.
SCNN generated 1,000,000 adversarial hard samples in 50 cycles. For comparison with random augmentation, we use the Augmentor tool~\cite{bloice2017augmentor} that applies random augmentation operations, such as rotation, scaling, shearing, flipping, random erasing, and elastic transformations. The results based on the online strategy for baseline model, Augmentor tool, and SCNN are shown in Fig.~\ref{fig1}.  It can be observed that the SCNN approach experiences less overfitting compared to other approaches. The final accuracy results are depicted in Table~\ref{tbl2}. In Fig.~\ref{fig2}, we illustrate the process of synthesizing a hard sample by applying gradient descent on the trainable parameters of the transformer, where a  sample with true label 2 is transformed, such that its label is predicted as 8.

\subsection{Fashion MNIST}
\begin{figure*}[htbp]
	\centerline{\includegraphics[width=\textwidth]{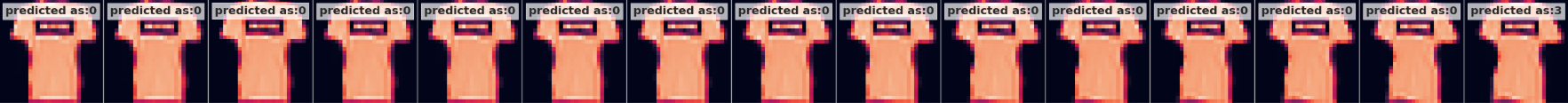}}
	\caption{Data manipulation based on differentiable grid on Fashion-MNIST sample data}
	\label{fig5}
\end{figure*}

We use the entire 60,000 samples in the training dataset of Fashion MNIST, and generated 600,000 adversarial hard samples in 10 cycles.
Once the baseline model was trained by the primary data, and another time, it is trained by the adversarial and primary data. The results are shown in Fig.~\ref{fig4}, and the final accuracy after 100 epochs is shown in Table~\ref{tbl3}. It can be observed that SCNN experiences less overfitting compared to the baseline model. Fig.~\ref{fig5} illustrates the synthesis of a hard sample by applying gradient descent on the trainable parameters of  the differentiable grid, where the goal is to manipulate a sample with true label 0 (T-shirt), such that it is predicted as label 3 (Dress).

\subsection{Cifar10}
\begin{figure*}[htbp]
	\centerline{\includegraphics[width=\textwidth]{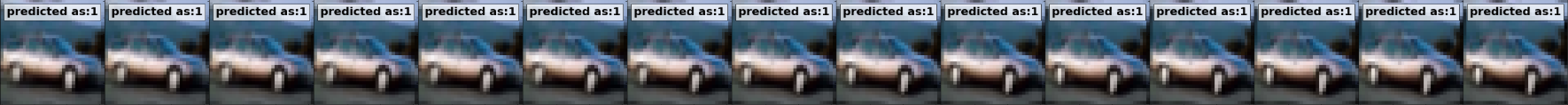}}
	\caption{Data manipulation based on differentiable grid on Cifar10 sample data}
	\label{fig6}
\end{figure*}

\begin{figure*}[h!]
	\centering
	\begin{subfigure}[t]{0.42\textwidth}
		\centering
		\centerline{\includegraphics[width=\columnwidth]{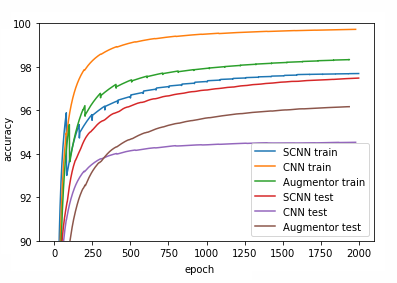}}

		\caption{Online Strategy for limited MNIST dataset}
		\label{fig1}
	\end{subfigure} \hspace{0.05\textwidth} %
	~ 
	\begin{subfigure}[t]{0.42\textwidth}
	\centering
		\centerline{\includegraphics[width=0.95\columnwidth]{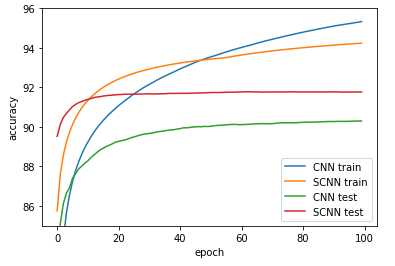}}
	\caption{Offline Strategy for Fashion-MNIST dataset}
	\label{fig4}
	\end{subfigure}%
	~ 
	\begin{subfigure}[t]{0.42\textwidth}
	\centering
	\centerline{\includegraphics[width=0.95\columnwidth]{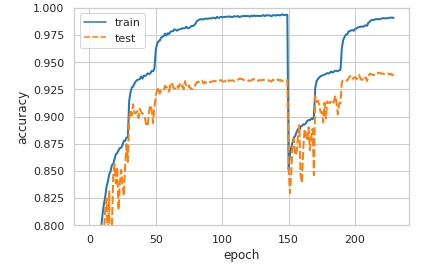}}
	\caption{Offline Strategy for Cifar10 dataset}
	\label{fig7}
	\end{subfigure}
	\caption{Performance of SCNN for different datasets}
\end{figure*}

This dataset is more complicated, consisting of 50,000 training data. We generated 100,000 adversarial data in 5 cycles. The baseline model was trained by 1.the primary data, 2.the primary data with random augmentation, 3.the primary data with AutoAugment approach which is introduced in \cite{AutoAugment2019} and 4.the primary data with the SCNN. The results are shown in Fig.~\ref{fig7}, and the final accuracy after 100 epochs is shown in Table~\ref{tbl4}. Fig.~\ref{fig6}  illustrates synthesizing a hard sample by applying gradient descent on the trainable parameters of the differentiable grid, where the generated sample is finally predicted correctly, but with less confidence. So we can see that after multiple cycles, the label of the adversarial data may still be predicted correctly.

\begin{table}[t!]
	\caption{Test accuracy of different approaches on the Limited MINST(1000 train data)}
	\label{tbl2}
	\centering
	\begin{tabular}{lll}
		\toprule
		Algorithms     & test accuracy(online strategy) & test accuracy(offline strategy)   \\
		\midrule
		Baseline   &   94.26 &  94.26     \\ 
		Baseline + Augmentor &    96.2 &  96.64       \\ 
		Baseline + SCNN   &   97.63  &   \textbf{98.25}    \\ 
		\bottomrule
	\end{tabular}
\end{table}

\begin{table}[t!]
	\caption{Test accuracy of CNN baseline model and SCNN on the Fashion MNIST dataset}
	\label{tbl3}
	\centering
	\begin{tabular}{lll}
		\toprule
		Network Type     & test accuracy(online strategy) & test accuracy(offline strategy)   \\
		\midrule
		Baseline CNN   &   90.11 &  90.11     \\ 
		Baseline + Augmentor &    90.54 &  90.74      \\
		SCNN   &   91.03  &   \textbf{91.45}    \\ 
		\bottomrule
	\end{tabular}
\end{table}

\begin{table}[t!]
	\caption{Test accuracy of on the Cifar10 dataset}
	\label{tbl4}
	\centering
	\begin{tabular}{lll}
		\toprule
		Network Type     & test accuracy    \\
		\midrule
		Resnet18   &   89.2     \\ 
		Resnet18 + Random Augmentation   &   93.3     \\ 
		Resnet18 + AutoAugment   &   93.83     \\ 
		Resnet18 + SCNN   &   \textbf{94.11}     \\ 
		\bottomrule
	\end{tabular}
\end{table}

%% file: Conclusion.tex
\section{Conclusion and Future Work}
We introduced a new approach for generating harder samples for a deep neural network. An SCNN contains two sub-networks, an embedded network, and a synthesizer network. We also proposed three differentiable image manipulator components which parameters can be learned in a way that can transfer a sample to the boundary of its true label and its nearest wrong label. We observed that the accuracy and the robustness of SCNNs are improved in comparison with their baseline models. 
As a future work, this approach can be extended by designing differentiable manipulator components for other types of input data, such as texts, graphs, and voice.